\documentclass[letterpaper]{article}
\usepackage[preprint]{aaai2027}
\usepackage[hyphens]{url}
\usepackage{graphicx}
\usepackage{amsmath}
\usepackage{amssymb}
\usepackage{bm}
\usepackage{natbib}
\usepackage{caption}
\usepackage{float}
\usepackage{algorithm}
\usepackage{algorithmic}
\usepackage{newfloat}
\usepackage{listings}
\usepackage{booktabs}
\usepackage{multirow}
\usepackage{xcolor}

\title{AdaDINO: Pair-Aware In-Backbone Adaptation of Frozen DINO \\ for Efficient Remote Sensing Change Detection}
\author{
Xu Zhang\textsuperscript{\rm 1},
Xinqing Li\textsuperscript{\rm 1},
Jianpeng Xie\textsuperscript{\rm 1},
Zeshuai Zhu\textsuperscript{\rm 1},
Xin He\textsuperscript{\rm 2},
Yun Liu\textsuperscript{\rm 1,3,4}\thanks{Corresponding author: Yun Liu (liuyun@nankai.edu.cn)}
}

\affiliations{
\textsuperscript{\rm 1}VCIP, College of Computer Science, Nankai University\\
\textsuperscript{\rm 2}School of Computer Science and Engineering, Tianjin University of Technology\\
\textsuperscript{\rm 3}Academy for Advanced Interdisciplinary Studies, Nankai University\\
\textsuperscript{\rm 4}Nankai International Advanced Research Institute, Shenzhen Futian
}

\begin{document}

\maketitle

\begin{abstract}
Vision foundation models (VFMs) such as DINO are pretrained for single-image representation, whereas remote sensing change detection requires reasoning over a bi-temporal pair. Existing VFM-based methods usually encode the two images independently and compare them only afterward, leaving the VFM backbone unaware of cross-temporal relations. To bridge this mismatch, we present \textbf{AdaDINO}, a pair-aware in-backbone adaptation framework that equips a frozen DINO encoder with bi-temporal interaction for efficient change detection. Its core component, \textbf{Change-aware Gated Local Adaptation (CGLA)}, couples the two streams after selected frozen blocks and injects a shared temporal residual into them with opposite signs, enhancing genuine change responses while preserving the pair midpoint. \textbf{Batch-Shared Chunk Selection (BSCS)} further reduces feed-forward network (FFN) computation by retaining a batch-shared subset of channel chunks that can be executed as a compact dense FFN. A \textbf{CGLA-Prior-Guided Refinement (CPGR)} decoder reuses encoder-side change responses for coarse-to-fine prediction. Experiments on four remote sensing change detection benchmarks show that AdaDINO achieves competitive or superior performance against VFM-based baselines, with the largest gain on the category-agnostic SYSU-CD dataset. With 62.5\% of the FFN hidden width removed, AdaDINO still achieves an F1 score of 85.29\% on SYSU-CD while delivering a $1.41\times$ throughput speedup. The code will be released.
\end{abstract}

\begin{figure}[!t]
  \centering
  \includegraphics[width=\columnwidth]{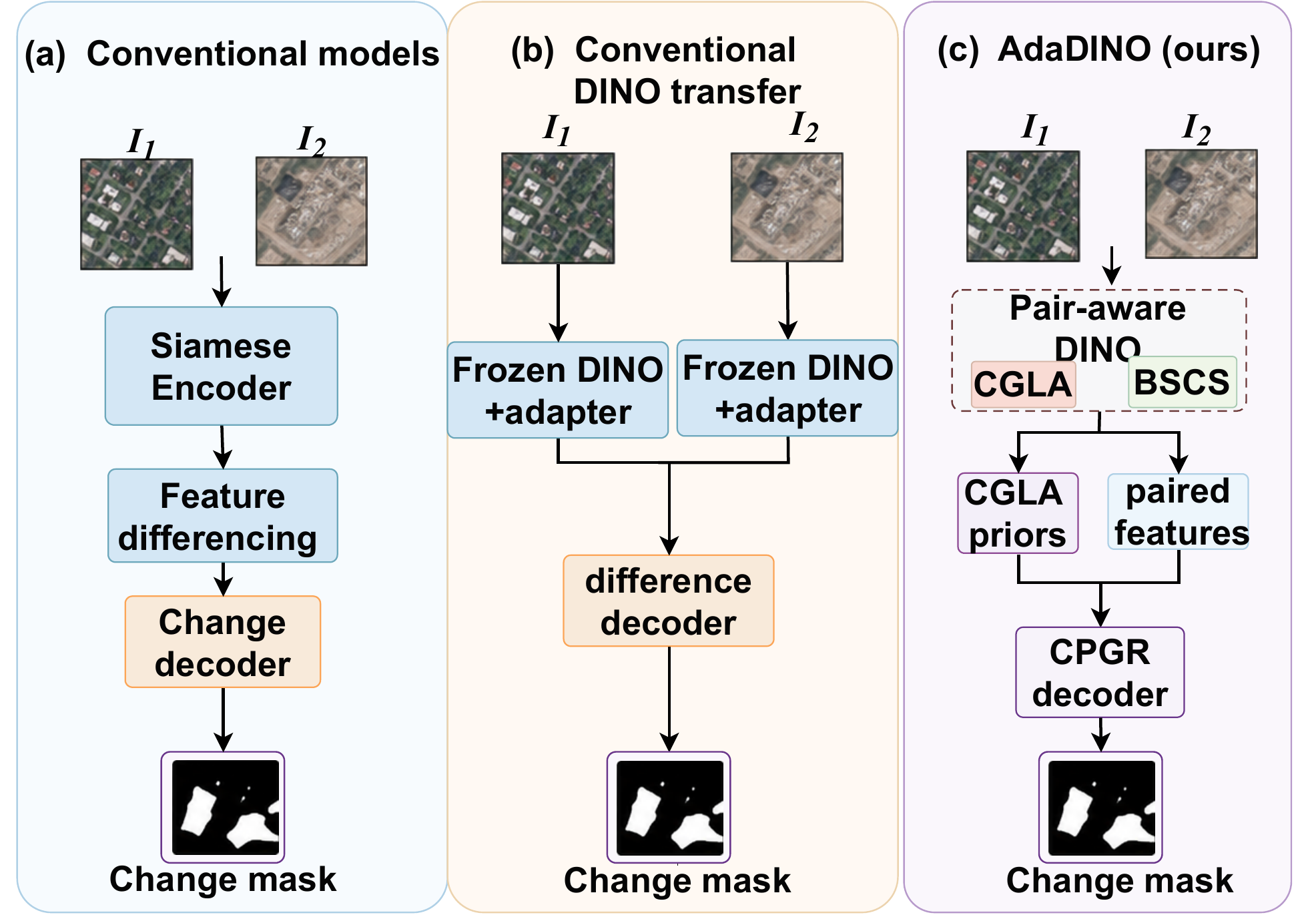}
  \captionsetup{skip=2pt}
    \caption{Comparison of bi-temporal change-detection paradigms.
    (a) Conventional Siamese detection.
    (b) Conventional DINO transfer with independent stream adaptation.
    (c) AdaDINO with pair-aware in-backbone adaptation.}
      \label{fig:teaser}
  \vspace{-6mm}
\end{figure}

\section{Introduction}
Remote sensing change detection aims to locate meaningful changes between two co-registered images of the same scene taken at different times. Change detection is challenging because genuine changes are often sparse, whereas most scene content remains unchanged. Meanwhile, variations in illumination, season, and imaging conditions can introduce pseudo-changes that are visually more salient than the changes of interest~\cite{hussain2013change,chen2020spatial,feaspect}. Consequently, meaningful change evidence forms a sparse signal within a highly redundant image pair, requiring the model to directly capture bi-temporal interactions, suppress nuisance variation, and allocate its capacity to change-relevant cues.

The modeling of bi-temporal image pairs has progressed from post-hoc feature comparison toward more explicit cross-temporal interaction. Early Siamese convolutional neural networks (CNNs) process the two images with weight-shared encoders and compare the resulting features mainly through decoder-side fusion or differencing~\cite{fcsiam}. Transformer-based methods subsequently introduce long-range spatial modeling and more explicit interaction between the two temporal streams \cite{bit,changeformer}. More recently, vision foundation models (VFMs) have provided stronger pretrained representations and have been transferred to change detection through task-specific bridging modules~\cite{ban}, differential decoders~\cite{changedino}, and parameter-efficient tuning strategies~\cite{peftcd,lora,vitadapter}. However, as illustrated in Fig.~\ref{fig:teaser}(b), many VFM-based approaches retain the image-wise operation of the pretrained backbone. Since the pretrained backbone is designed for single-image representation learning, the two observations are still encoded largely independently within the backbone: adaptation is performed largely within each stream, while explicit pairwise interaction is deferred to external fusion modules or the decoder. Consequently, cross-temporal evidence cannot directly shape intermediate backbone representations, limiting the encoder's ability to suppress nuisance variations and preserve subtle change cues before decoding.

The key limitation is therefore not the representation quality of DINO itself, but where and how the two temporal observations interact. Under independent encoding, subtle changes must survive two separate feature hierarchies before comparison, while nuisance variations cannot be jointly suppressed during feature formation. The highly redundant pair structure also suggests an opportunity for more selective computation: because most scene content remains unchanged, executing every feed-forward network (FFN) at its full hidden width may devote substantial computation to pair-irrelevant responses. Existing dynamic vision transformers, however, mainly rely on token- or sample-wise routing, whose irregular sparsity is difficult to translate into practical speedups. They also do not address pair-conditioned FFN channel selection while preserving regular batched execution~\cite{adavit,dynamicvit,avit}. An effective solution should therefore make the frozen encoder pair-aware while reducing FFN computation in a structured and hardware-friendly manner.

To this end, we present \textbf{AdaDINO} (shown in Fig.~\ref{fig:teaser}), a pair-aware in-backbone adaptation framework that moves bi-temporal reasoning into the frozen DINO encoder while making its FFN computation responsive to the input batch. \textbf{Change-aware Gated Local Adaptation (CGLA)} is attached after selected frozen blocks, where it derives pair-conditioned change cues and injects an antisymmetric temporal residual into the two streams, strengthening change-sensitive representations without shifting their midpoint. \textbf{Batch-Shared Chunk Selection (BSCS)} then uses pooled pair statistics to select a batch-shared top-$k$ set of contiguous FFN chunks, allowing the retained computation to be executed as a compact dense FFN at inference. Finally, \textbf{CGLA-Prior-Guided Refinement (CPGR)} converts the multi-depth CGLA responses into stage-aligned priors for coarse-to-fine decoding, directly carrying encoder-side change evidence into mask refinement instead of relying solely on decoder-side bi-temporal differences.

On four benchmarks AdaDINO is competitive with or better than VFM-based baselines, and on the category-agnostic SYSU-CD~\cite{sysu} it stays the most accurate even after $62.5\%$ of the FFN hidden width is removed.

Our main contributions are summarized as follows:
\begin{itemize}
  \item We introduce Change-aware Gated Local Adaptation (CGLA), which couples the two temporal streams after selected frozen DINO blocks through pair-conditioned antisymmetric residuals, enabling in-backbone change reasoning while keeping the pretrained parameters fixed.
  \item We propose Batch-Shared Chunk Selection (BSCS), which selects a batch-shared subset of contiguous FFN chunks from pooled pair statistics, enabling input-adaptive width reduction with compact dense execution. We further find that the CGLA-adapted backbone withstands this pruning  far better than an unadapted one.
  \item We develop CGLA-Prior-Guided Refinement (CPGR), which propagates multi-depth encoder-side change responses as stage-aligned priors, guiding coarse-to-fine mask refinement beyond conventional decoder-side bi-temporal differences.
\end{itemize}
\begin{figure*}[!t]
  \centering
  \includegraphics[width=\textwidth]{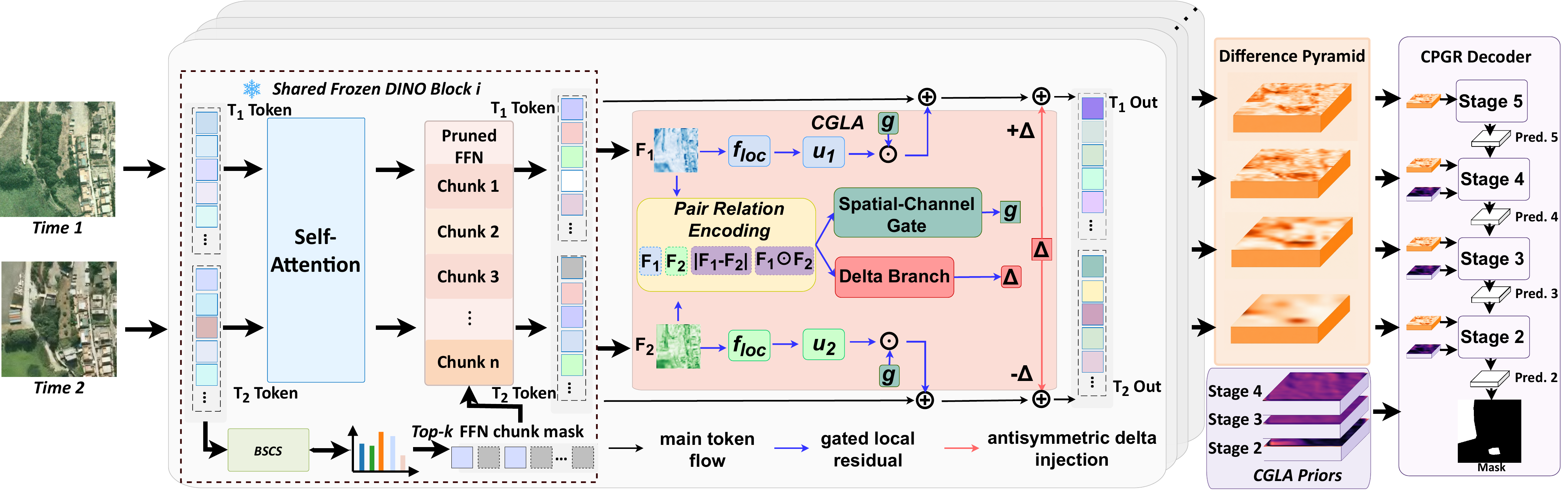}
  \captionsetup{skip=2pt}
    \caption{Overview of AdaDINO. \emph{Left}: a shared frozen DINOv3 block with BSCS chunk selection. \emph{Middle}: CGLA, which couples the two streams and caches its change responses. \emph{Right}: the CPGR decoder guided by stage-aligned CGLA priors. Arrow styles distinguish the main token flow, the gated local residual, and the antisymmetric delta injection. }
  \label{fig:overview}
  \vspace{-4mm}
\end{figure*}

\section{Related Work}
\subsection{Bi-temporal Encoding and Interaction}
Modern change detectors commonly adopt Siamese architectures, processing co-registered images with weight-shared encoders and constructing change evidence through feature comparison or fusion~\cite{fcsiam,bit,changeformer}. CNN-based models emphasize local patterns~\cite{fcsiam,snunet}, whereas transformer-based approaches~\cite{bit,changeformer} capture longer-range spatial and temporal dependencies. In many such designs, however, explicit interaction occurs mainly after feature extraction, leaving intermediate representations weakly conditioned on cross-temporal relations. Recent methods therefore move interaction into the encoder: MetaChanger~\cite{changer} inserts feature-exchange layers between the two streams, while SARAS-Net~\cite{sarasnet} applies relation-aware cross-attention before feature subtraction. AdaDINO extends this direction to VFMs through CGLA, which injects a pair-conditioned temporal residual into the two streams with opposite signs after selected DINO blocks, reshaping their cross-temporal discrepancy while preserving the pair midpoint.

\subsection{Vision Foundation Models for Change Detection}
VFMs provide transferable representations for change detection, but existing methods differ in where bi-temporal interaction occurs. BAN \cite{ban} keeps the backbone frozen and injects aligned pretrained features into a task-specific branch, leaving explicit interaction outside the backbone. ChangeDINO \cite{changedino} combines frozen DINOv3 features with a lightweight convolutional stream and performs comparison in a differential decoder, while PeftCD \cite{peftcd} inserts LoRA \cite{lora} adapters into a shared VFM encoder before cross-temporal exchange and lightweight decoding. TTP \cite{ttp} brings interaction into VFM encoding through time-traveling gates in a SAM backbone \cite{sam}. GenCD~\cite{gencd} inserts a low-rank exchange branch into the QKV projections of a frozen VFM, adapting each stream with a compressed view of the other. The proposed AdaDINO instead injects a shared pair-conditioned residual with opposite signs after selected frozen DINO blocks, reshaping the cross-temporal discrepancy while preserving the pair midpoint. This design addresses the mismatch between single-image VFM pretraining and pairwise change reasoning.

\subsection{Efficient Change Detection}
Efficient change detection has been explored through compact architectures with lightweight backbones and fusion modules~\cite{lcdnet}, as well as static filter pruning of trained detectors~\cite{oaip}. Although effective for resource-constrained deployment~\cite{oaip}, these approaches fix the network structure before inference and apply the same computation to every bi-temporal pair. Input-adaptive vision transformers instead reduce computation through token pruning, layer skipping, early exiting, or gating~\cite{adavit,dynamicvit,avit,tome,sparsevit}, while related remote sensing methods route tokens to exploit spatial redundancy~\cite{dynamicvis}. However, token- or sample-specific sparsity often complicates regular batched execution. Parameter-efficient methods reduce trainable parameters \cite{lora,vitadapter,lorand,peftvit}, but do not directly reduce dense FFN inference cost.

Structured MLP pruning assigns fixed per-block FFN widths based on neuron importance \cite{amp}. In contrast, the proposed BSCS predicts pair-conditioned importance for contiguous FFN chunks and selects a batch-shared top-$k$ subset. This design adapts the selected computation to each input batch while retaining compact dense execution and preserving all spatial tokens for change-mask prediction.

\section{Method}
\label{sec:method}
\subsection{Framework Overview}
\label{sec:overview}

Given a co-registered bi-temporal image pair $(I_1,I_2)$, our goal is to predict a binary change map matching the ground-truth annotation $Y\in\{0,1\}^{H\times W}$. We employ a shared frozen DINOv3~\cite{dinov3} backbone as the Siamese encoder. Let $E(\cdot)$ denote the patch-embedding operation and $B^l(\cdot)$ denote the $l$-th frozen Transformer block. The blockwise encoding process is written as
\begin{equation}
    \mathbf{x}_i^0 = E(I_i), \tilde{\mathbf{x}}_i^l = B^l\!\left(\mathbf{x}_i^{l-1}\right),
     i\in\{1,2\}, l=1,\dots,L,
    \label{eq:backbone-encoding}
\end{equation}
where $L$ is the number of backbone blocks. In a vanilla Siamese configuration, $\mathbf{x}_i^l=\tilde{\mathbf{x}}_i^l$, and the two temporal streams are processed independently throughout the backbone. Explicit change reasoning is therefore deferred to subsequent feature-comparison and decoding stages. Consequently, cross-temporal evidence cannot directly influence the formation of intermediate representations, while every FFN is
executed at its full hidden width for each input pair.

AdaDINO comprises three complementary components (Fig.~\ref{fig:overview}). CGLA couples the two temporal streams after selected frozen blocks, allowing pair-conditioned residual updates to reshape subsequent backbone representations. It also produces multi-depth change responses that can be reused during decoding. BSCS predicts FFN chunk importance from the paired representations and executes a batch-shared top-$k$ subset at inference. Finally, CPGR aggregates the CGLA responses into stage-aligned priors to guide coarse-to-fine change-mask decoding.

\subsection{Change-aware Gated Local Adaptation}
\label{sec:cgla}
DINO processes $I_1$ and $I_2$ independently, so its intermediate representations are not explicitly conditioned on their cross-temporal relations. To address this limitation, CGLA is inserted after selected frozen DINO blocks indexed by $\mathcal{L}_c$. At each selected depth, the original DINO block is executed unchanged, after which CGLA applies a lightweight pair-conditioned residual update to the paired token states before they enter the subsequent block. All pretrained DINO parameters remain frozen throughout.
 
For an adapted depth $l\in\mathcal{L}_c$, let $\tilde{\mathbf{x}}_i^l$ denote the output of the $l$-th frozen DINO block for stream $i\in\{1,2\}$. We apply layer normalization to $\tilde{\mathbf{x}}_i^l$, extract its patch tokens, and reshape them into a spatial feature map $\mathbf{F}_i^l$. Prefix tokens, including the class and register tokens, are excluded from this spatial processing and remain unchanged by CGLA. The two appearance features are then combined with their absolute discrepancy and multiplicative agreement to construct a pairwise relation descriptor:
\begin{equation}
  \mathbf{z}^l =
  [\,\mathbf{F}_1^l,\mathbf{F}_2^l,\mathbf{D}^l,\mathbf{A}^l\,],
  \quad
  \mathbf{c}^l =
  f_{\mathrm{ctx}}\!\left(f_{\mathrm{rel}}(\mathbf{z}^l)\right),
  \label{eq:cgla-context}
\end{equation}
where $[\cdot]$ denotes channel-wise concatenation, $\mathbf{D}^l=|\mathbf{F}_1^l-\mathbf{F}_2^l|$, and $\mathbf{A}^l=\mathbf{F}_1^l\odot\mathbf{F}_2^l$. The lightweight module $f_{\mathrm{rel}}$ projects the concatenated relation cues into a compact hidden representation, while $f_{\mathrm{ctx}}$ further captures local spatial context to produce the change-context feature $\mathbf{c}^l$. This process corresponds to the \emph{pair relation encoding} block in Fig.~\ref{fig:overview}.
 
From the change-context feature $\mathbf{c}^l$, CGLA derives a spatial gate and a channel gate, whose broadcasted product forms a joint local gate:
\begin{equation}
  \mathbf{g}_s^l =
  \sigma\!\left(\mathbf{W}_s\mathbf{c}^l\right), 
  \mathbf{g}_c^l =
  \sigma\!\left(
  \mathbf{W}_c\operatorname{GAP}(\mathbf{c}^l)
  \right), 
  \mathbf{g}^l =
  \mathbf{g}_s^l\odot\mathbf{g}_c^l .
  \label{eq:cgla-gate}
\end{equation}
Here, $\mathbf{W}_s$ and $\mathbf{W}_c$ denote the $1\times1$ convolutions used by the spatial and channel gating branches, respectively. The sigmoid function $\sigma(\cdot)$ bounds the gate values, and $\operatorname{GAP}(\cdot)$ denotes global average pooling over the spatial dimensions. The spatial gate $\mathbf{g}_s^l$ captures location-dependent responses and the channel gate $\mathbf{g}_c^l$ provides channel-wise modulation.

A gated delta branch then produces a pair-conditioned temporal update:
\begin{equation}
  \mathbf{d}^l =
  \mathbf{g}^l\odot f_{\Delta}(\mathbf{c}^l), \quad 
  \bm{\Delta}^l =
  \bm{\gamma}_{\Delta}\odot\mathbf{d}^l ,
  \label{eq:cgla-delta}
\end{equation}
where $f_{\Delta}$ denotes the convolutional delta branch and $\bm{\gamma}_{\Delta}$ is a learnable channel-wise scaling vector, broadcast over the spatial dimensions.  In parallel, a shared local operator $f_{\mathrm{loc}}$ is applied separately to the two streams, yielding stream-specific local updates
\begin{equation}
\mathbf{u}_i^l =
f_{\mathrm{loc}}(\mathbf{F}_i^l), \quad
i\in\{1,2\}.
\end{equation}
The gated local and temporal updates are fused on the spatial grid and then written back to the patch-token positions:
\begin{equation}
\begin{gathered}
  \mathbf{r}_i^l =
  \eta\left(
  \mathbf{g}^l\odot\bm{\gamma}_{\mathrm{self}}
  \odot\mathbf{u}_i^l
  + s_i\bm{\Delta}^l
  \right),\\
  \mathbf{x}_i^l =
  \tilde{\mathbf{x}}_i^l +
  \operatorname{Tok}\!\left(\mathbf{r}_i^l\right).
\end{gathered}
\label{eq:cgla-update}
\end{equation}
Here, $s_1=1$ and $s_2=-1$ in the antisymmetric mode, $\eta$ is a learnable residual scale, and
$\bm{\gamma}_{\mathrm{self}}$ is a learnable channel-wise scale for the local updates. The operator $\operatorname{Tok}(\cdot)$ maps a spatial residual back to the corresponding patch-token positions, while leaving the prefix tokens unchanged. Because the temporal residual is injected with opposite signs, its contribution cancels in the midpoint of the bi-temporal feature pair and contributes $2\eta\bm{\Delta}^l$ to their cross-temporal difference. 

The stream-specific local updates are not constrained to be antisymmetric, allowing each temporal stream to retain its own local refinement. Beyond updating the token representations, CGLA also exposes its intermediate change-aware responses as spatial priors for decoding. At each adapted depth, we record
\begin{equation}
  \mathbf{q}_s^l = \mathbf{g}_s^l, \quad
  \mathbf{q}_{\mathrm{loc}}^l =
  \operatorname{mean}_c\!\left|\mathbf{g}^l\right|, \quad
  \mathbf{q}_{\Delta}^l =
  \operatorname{mean}_c\!\left|\mathbf{d}^l\right|,
  \label{eq:cgla-priors}
\end{equation}
where $\operatorname{mean}_c(\cdot)$ denotes averaging over the channel dimension. These response maps are direct byproducts of CGLA and therefore require no additional feature-extraction pass. In the main configuration, CPGR uses the spatial response $\mathbf{q}_s^l$ and the delta response $\mathbf{q}_{\Delta}^l$, whereas $\mathbf{q}_{\mathrm{loc}}^l$ is retained to support alternative prior configurations. CGLA thus generates change-aware responses during backbone encoding and makes them directly available to the decoder.

 \begin{figure*}[!t]
  \centering
  \includegraphics[width=\textwidth]{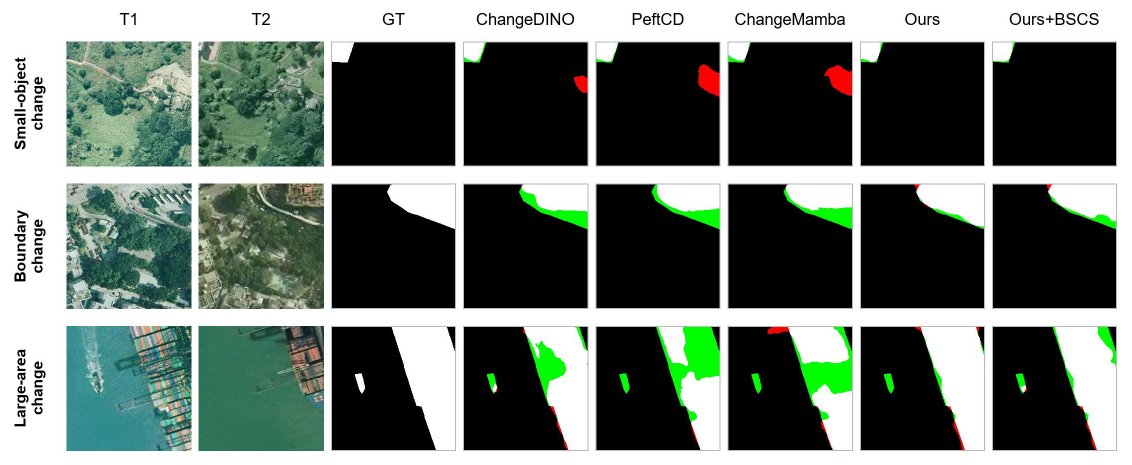}
  \captionsetup{skip=2pt}
  \caption{Qualitative comparison on SYSU-CD. Rows show different change regimes: small-object, boundary-sensitive,
and large-area changes. Prediction maps use white/black/red/green for TP/TN/FP/FN, respectively. ``Ours+BSCS'' retains
6 of 16 FFN chunks ($37.5\%$) and preserves predictions close to the full model while reducing computation.}
  \label{fig:qualitative}
  \vspace{-4mm}
\end{figure*}

\subsection{Batch-Shared Chunk Selection}
\label{sec:pruning}
A standard DINO block executes its feed-forward network at the full hidden width for every bi-temporal input, although different image pairs and encoder depths may not require the same FFN channels. To reduce this unnecessary computation, BSCS partitions the FFN hidden dimension into
structured chunks and employs a lightweight block-specific policy to select an input-dependent subset for execution. The selection is shared across all samples in a batch, allowing the retained chunks to be assembled into compact dense projections for efficient inference.

We divide the FFN hidden dimension into $C$ contiguous chunks. Let $\mathbf{x}_{i,b}^{l-1}$ denote the token sequence of stream $i$ for sample $b$ entering the $l$-th block. Before the FFN is executed, a lightweight block-specific policy network summarizes the
paired token streams and predicts chunk scores $\mathbf{s}_b^l\in\mathbb{R}^{C}$:
\begin{equation}
  \mathbf{h}_b^l =
  \operatorname{PoolPair}\!\left(
  \mathbf{x}_{1,b}^{l-1},
  \mathbf{x}_{2,b}^{l-1}
  \right), \quad
  \mathbf{s}_b^l =
  \pi^l\!\left(\mathbf{h}_b^l\right),
  \label{eq:policy-score}
\end{equation}
where $\pi^l$ denotes the policy network associated with the $l$-th block. After excluding the prefix tokens, $\operatorname{PoolPair}(\cdot)$ concatenates the mean-pooled features of the two streams, their mean absolute token-wise difference, and their mean token-wise multiplicative
agreement. The resulting score is therefore conditioned jointly on the bi-temporal pair rather than on either observation alone. 

To obtain a batch-shared selection, we first average the per-sample chunk logits within the batch and convert the aggregated logits into soft selection probabilities:
\begin{equation}
  \bar{\mathbf{s}}^l =
  \frac{1}{B}\sum_{b=1}^{B}\mathbf{s}_b^l, \quad
  \mathbf{p}^l =
  \sigma\!\left(\bar{\mathbf{s}}^l/\tau\right),
  \label{eq:batch-score}
\end{equation}
where $\tau$ is a temperature hyperparameter controlling the sharpness of the probabilities. We then retain the $k$ highest-scoring chunks:
\begin{equation}
  \mathbf{m}_{\mathrm{hard}}^l =
  \operatorname{TopK}\!\left(\mathbf{p}^l,k\right),
  \label{eq:batch-shared-topk}
\end{equation}
where $\operatorname{TopK}(\cdot,k)$ returns a $k$-hot binary mask $\mathbf{m}_{\mathrm{hard}}^l\in\{0,1\}^{C}$. All samples in the batch share the same mask.
Because the top-$k$ operation is non-differentiable, we use a straight-through estimator:
\begin{equation}
  \tilde{\mathbf{m}}^l =
  \operatorname{sg}\!\left(
  \mathbf{m}_{\mathrm{hard}}^l-\mathbf{p}^l
  \right)
  + \mathbf{p}^l,
  \label{eq:ste-mask}
\end{equation}
where $\operatorname{sg}(\cdot)$ denotes stop-gradient. Thus, the forward pass follows the discrete top-$k$ decision, while the backward pass propagates gradients through the soft probabilities $\mathbf{p}^l$.

Let the FFN of the $l$-th block be written as $\operatorname{FFN}^l(\mathbf{x})
=\mathbf{W}_2^l\rho(\mathbf{W}_1^l\mathbf{x})$, where $\mathbf{W}_1^l$ and $\mathbf{W}_2^l$ denote the input and output projections, respectively, and $\rho(\cdot)$ is the activation function. For clarity, biases and dropout are omitted from the notation. During training, the full hidden representation is first computed, after which the chunk-level mask is expanded over the corresponding hidden channels:
\begin{equation}
  \operatorname{FFN}_{\tilde{\mathbf{m}}}^l(\mathbf{x}) =
  \mathbf{W}_2^l\!\left(
  \operatorname{Expand}(\tilde{\mathbf{m}}^l)
  \odot
  \rho(\mathbf{W}_1^l\mathbf{x})
  \right).
  \label{eq:masked-ffn}
\end{equation}
Here, $\operatorname{Expand}(\cdot)$ assigns each chunk-level gate to all hidden channels belonging to that chunk. Because both full-width projections are still evaluated, this masked training procedure does not reduce the training cost. We therefore keep all chunks active during an initial warm-up period and enable the learned selection afterward to stabilize optimization.

At inference time, the batch-shared binary mask allows the active rows of $\mathbf{W}_1^l$ and the corresponding active columns of $\mathbf{W}_2^l$ to be gathered into compact submatrices. The selected chunks can thus be evaluated using reduced-width dense projections, while the inactive chunks are skipped entirely. 

\subsection{CGLA-Prior-Guided Refinement}
\label{sec:decoder}
Conventional coarse-to-fine decoders derive refinement cues mainly from decoder features and coarser predictions. CPGR additionally reuses the multi-depth change-aware responses generated from CGLA, directly connecting pair-aware encoder adaptation with change-mask refinement.

The paired multi-scale features are first converted into a difference pyramid, $\mathbf{D}_r = \left|\mathbf{F}_{1,r}-\mathbf{F}_{2,r}\right|$, $r\in\{2,3,4,5\}$, where $\mathbf{F}_{i,r}$ denotes the feature of temporal stream $i$ at pyramid level $r$. The difference pyramid is then decoded from coarse to fine. To align the CGLA responses with this hierarchy, we partition the ordered adapted depths $\mathcal{L}_c$ into four consecutive groups $\{\mathcal{G}_r\}_{r=2}^{5}$ and aggregate the spatial and delta responses within each group:
\begin{equation}
  \mathbf{Q}_{r,s} =
  \frac{1}{|\mathcal{G}_r|}
  \sum_{l\in\mathcal{G}_r}\mathbf{q}_s^l,
  \quad
  \mathbf{Q}_{r,\Delta} =
  \frac{1}{|\mathcal{G}_r|}
  \sum_{l\in\mathcal{G}_r}\mathbf{q}_{\Delta}^l.
  \label{eq:stage-mean}
\end{equation}
This maps shallow-to-deep CGLA responses to four stage-aligned prior groups. Only the priors for $r\in\{2,3,4\}$ are injected into the decoder, since $P_5$ serves as the initial prediction and receives no prior guidance. At refinement stage $r$, the aligned responses are detached from the encoder-side computation graph, resized, concatenated, and projected through a zero-initialized $1\times1$ convolution:
\begin{equation}
  \mathbf{G}_r^{\mathrm{prior}} =
  \alpha_{\mathrm{prior}}\,
  \mathbf{W}_{\mathrm{prior}}
  \left[
    \mathcal{R}_r\!\left(
      \operatorname{sg}(\mathbf{Q}_{r,s})
    \right),
    \mathcal{R}_r\!\left(
      \operatorname{sg}(\mathbf{Q}_{r,\Delta})
    \right)
  \right],
  \label{eq:prior-gate}
\end{equation}
where $[\cdot]$ denotes channel-wise concatenation, $\mathcal{R}_r(\cdot)$ resizes a response map to the spatial resolution of stage $r$, and $\operatorname{sg}(\cdot)$ denotes stop-gradient. $\mathbf{W}_{\mathrm{prior}}$ is the weight of a zero-initialized $1\times1$ convolution without normalization or activation, and $\alpha_{\mathrm{prior}}$ is a learnable scaling factor. The encoder-derived prior is combined with guidance from the next coarser prediction:
\begin{equation}
  \mathbf{G}_r = \alpha_{\mathrm{pred}}\, f_{\mathrm{pred}}\!\left(\Psi\!\left(\operatorname{sg}(P_{r+1})\right) \right) + \mathbf{G}_r^{\mathrm{prior}},
  \label{eq:decoder-guide}
\end{equation}
where $P_{r+1}$ denotes the logits predicted at the coarser level $r+1$, $f_{\mathrm{pred}}$ is a lightweight projection, and $\alpha_{\mathrm{pred}}$ is a learnable scaling factor. $\Psi(\cdot)$ derives the change probability $p$, uncertainty $4p(1-p)$, and boundary response $|\nabla^2 p|$ from the coarse prediction. The resulting guidance modulates feature routing and fusion within each refinement stage. Predictions at all levels are supervised, and a lightweight morphological refiner is finally applied to the finest prediction to obtain the final change map $\hat{P}$.

\begin{table*}[!t]
  \centering
  \begin{minipage}[t]{0.69\textwidth}
    \vspace{0pt}
    \centering
\captionsetup{skip=2pt}
    \captionof{table}{
      Comparison on four change-detection benchmarks (F1/IoU, \%).
      Best and second-best results are shown in
      \textbf{bold} and \underline{underlined}, respectively.
      $\ddagger$ denotes our reproduction; ``--'' indicates results
      unavailable under a comparable split.
    }
    \label{tab:main}

    \resizebox{\linewidth}{!}{%
      \renewcommand{\arraystretch}{1.05}
      \setlength{\tabcolsep}{4.5pt}
      \begin{tabular}{@{}llcccccccc@{}}
        \toprule
        & &
        \multicolumn{2}{c}{SYSU-CD} &
        \multicolumn{2}{c}{WHU-CD} &
        \multicolumn{2}{c}{LEVIR-CD} &
        \multicolumn{2}{c}{LEVIR-CD+} \\
        \cmidrule(lr){3-4}
        \cmidrule(lr){5-6}
        \cmidrule(lr){7-8}
        \cmidrule(lr){9-10}
        Type & Method
        & F1 & IoU
        & F1 & IoU
        & F1 & IoU
        & F1 & IoU \\
        \midrule
        \multirow{4}{*}{CNN}
        & FC-EF
        & 75.81 & 61.04
        & 84.89 & 73.74
        & 88.90 & 80.03
        & 76.41 & 61.82 \\
        & FC-Siam-Diff
        & 75.79 & 61.01
        & 87.67 & 78.04
        & 89.69 & 81.31
        & 79.23 & 65.61 \\
        & FC-Siam-Conc
        & 75.18 & 60.23
        & 85.83 & 75.18
        & 89.89 & 81.64
        & 79.09 & 65.42 \\
        & SNUNet
        & 73.14 & 57.66
        & 87.70 & 78.09
        & 88.59 & 79.51
        & 78.56 & 64.70 \\
        \midrule
        \multirow{3}{*}{Transformer}
        & BIT
        & 79.62 & 66.14
        & 90.01 & 81.84
        & 90.03 & 81.87
        & 80.95 & 68.00 \\
        & ChangeFormer
        & -- & --
        & 90.30 & 82.32
        & 88.83 & 79.90
        & 77.54 & 63.31 \\
        & ChangeTitans
        & 83.24 & 71.18
        & -- & --
        & 91.52 & 84.36
        & 86.37 & 75.84 \\
        \midrule
        \multirow{4}{*}{Mamba}
        & ChangeMamba
        & 83.11 & 71.10
        & 92.55 & 86.13
        & 90.16 & 82.09
        & 82.68 & 70.47 \\
        & RS-Mamba
        & 78.99 & 65.28
        & 91.97 & 85.14
        & 90.82 & 83.18
        & 81.07 & 68.16 \\
        & CDMamba
        & 79.07 & 65.38
        & 91.10 & 83.65
        & 90.75 & 83.07
        & 81.77 & 69.16 \\
        & ST-Mamba
        & 83.39 & 71.50
        & 94.72 & 90.15
        & 91.92 & 85.05
        & 84.01 & 72.44 \\
        \midrule
        \multirow{6}{*}{VFM}
        & BAN
        & -- & --
        & 94.49 & 89.56
        & 91.70 & 84.67
        & -- & -- \\
        & FAEWNet
        & -- & --
        & 94.99 & 90.45
        & \textbf{92.41} & \textbf{85.89}
        & -- & -- \\
        & ChangeDINO
        & 84.70 & 73.46
        & 94.88$^\ddagger$ & 90.26$^\ddagger$
        & 92.31 & 85.72
        & \underline{86.99}$^\ddagger$ & \underline{76.98}$^\ddagger$ \\
        & PeftCD
        & 84.93 & 73.81
        & \textbf{95.86} & \textbf{92.05}
        & 92.25 & 85.62
        & 85.89$^\ddagger$ & 75.26$^\ddagger$ \\
        \cmidrule(lr){2-10}
        & Ours
        & \textbf{86.12} & \textbf{75.62}
        & \underline{95.30} &\underline{91.02}
        & \underline{92.35} &\underline{85.79}
        & \textbf{87.78} & \textbf{78.21} \\

        & Ours + BSCS
        & \underline{85.29} & \underline{74.35}
        & 94.31 & 89.24
        & 91.90 & 85.01
        & 86.60 & 76.36 \\
        \bottomrule
      \end{tabular}%
    }
  \end{minipage}%
  \hfill
 \begin{minipage}[t]{0.28\textwidth}
  \vspace{0pt}
  \centering

  \captionsetup{
    justification=raggedright,
    singlelinecheck=false,
    skip=15pt
  }

\captionof{table}{
  Ablation on SYSU-CD (F1/IoU, \%).
  Pair-blind uses matched insertion depths and a comparable
  trainable-parameter budget.
  Static Top-$6$ fixes training-set-derived masks; all $k{=}6$
  variants are retrained.
}
  \label{tab:ablation}

  \footnotesize
  \setlength{\tabcolsep}{3.5pt}
  \renewcommand{\arraystretch}{1.10}

  \begin{tabular*}{\linewidth}{
    @{\extracolsep{\fill}}lcc@{}
  }
    \toprule
    Configuration & F1 & IoU \\
    \midrule
    Frozen baseline
    & 83.77 & 72.07 \\
    \quad + Pair-blind
    & 84.64 & 73.37 \\
    \quad + CGLA
    & 85.61 & 74.84 \\
    \quad + CGLA + CPGR
    & 86.12 & 75.62 \\

    \midrule
    \multicolumn{3}{@{}l}{\emph{CGLA design}} \\
    \quad w/o $\Delta$ injection
    & 85.18 & 74.19 \\
    \quad symmetric $\Delta$
    & 85.48 & 74.65 \\

    \midrule
    \multicolumn{3}{@{}l}{\emph{FFN selection ($k{=}6$)}} \\
    \quad BSCS Top-$6$
    & 85.29 & 74.35 \\
    \quad Static Top-$6$
    & 84.58 & 73.28 \\
    \quad Random-$6$
    & 83.26 & 71.32 \\
    \quad Bottom-$6$
    & 82.77 & 70.61 \\
    \bottomrule
  \end{tabular*}
\end{minipage}
\vspace{-4mm}
\end{table*}

\section{Experiments}
\label{sec:exp}
 
\subsection{Experimental Setup}
\label{sec:setup}
 
\paragraph{Datasets.}
We evaluate on the category-agnostic SYSU-CD~\cite{sysu} and three building-change benchmarks: WHU-CD~\cite{whu},
LEVIR-CD~\cite{chen2020spatial}, and LEVIR-CD+~\cite{s2looking}. This benchmark suite evaluates complementary forms of generalization: SYSU-CD tests pair-aware adaptation across diverse change categories, while WHU-CD, LEVIR-CD, and LEVIR-CD+ examine whether the same design remains effective across established building-change benchmarks. Images are cropped into non-overlapping $256\times256$ pairs. Following common splits~\cite{cdmamba,stmamba}, SYSU-CD, WHU-CD, and LEVIR-CD contain $12{,}000/4{,}000/4{,}000$, $6{,}096/762/762$, and $7{,}120/1{,}024/2{,}048$ train/validation/test pairs, respectively; LEVIR-CD+ contains $10{,}192/5{,}568$ train/test pairs. SYSU-CD and LEVIR-CD use their official splits.
Following common practice, we report F1 score and Intersection-over-Union (IoU) on the change class as our primary metrics.
 
\paragraph{Compared methods.}
  We compare against representative methods spanning four families:
  CNN-based Siamese networks (FC-EF, FC-Siam-Diff, FC-Siam-Conc~\cite{fcsiam},  SNUNet~\cite{snunet}), transformer detectors (BIT~\cite{bit}, ChangeFormer~\cite{changeformer}, ChangeTitans~\cite{changetitans}), Mamba/state-space models (ChangeMamba~\cite{mambacd},  RS-Mamba~\cite{rsmamba}, CDMamba~\cite{cdmamba}, ST-Mamba~\cite{stmamba}), and VFM-based detectors (BAN~\cite{ban}, FAEWNet~\cite{faewnet}, ChangeDINO~\cite{changedino}, PeftCD~\cite{peftcd}). For baselines
  evaluated under the same protocol we quote published numbers: most are
  taken from~\cite{stmamba, cdmamba}, which share our splits; ChangeTitans is quoted from its own paper, while BAN and FAEWNet are from~\cite{faewnet}. 
\paragraph{Implementation details.}
CGLA is inserted at eight uniformly spaced depths,
$\mathcal{L}_c=\{2,5,8,11,14,17,20,23\}$.
Each FFN is divided into $C=16$ chunks. BSCS uses
$\tau=1$ and is enabled after a three-epoch all-chunk warm-up.
We use a frozen DINOv3 ViT-L/16 pretrained on SAT-493M
\cite{dinov3}, training only CGLA, BSCS, CPGR, and the decoder.
Inputs are cropped to $256\times256$ and resized to
$512\times512$ before entering the backbone.
We train for $100$ epochs with AdamW
($\mathrm{lr}=5\times10^{-4}$, weight decay $5\times10^{-4}$),
cosine decay to $10^{-7}$, and an effective batch size of $16$.
The objective combines focal loss ($\alpha=0.25$, $\gamma=4$)
and Dice loss over the final and four pyramid predictions after
upsampling, plus an auxiliary term on the CGLA responses.
The exact weighting is provided in Appendix~A.
We apply temporal swapping, flips, right-angle rotations, color
jitter, and random resized cropping ($p=0.5$ each), without
test-time augmentation.

\subsection{Main Results}
\label{sec:main}
 
  Table~\ref{tab:main} reports F1 and IoU on the four benchmarks. Unless otherwise  noted, ``Ours\,+\,BSCS'' retains $6$ of the $16$ feed-forward chunks  ($37.5\%$). AdaDINO attains the best F1 on SYSU-CD ($86.12\%$, 1.19 F1
points above the strongest baseline) and on LEVIR-CD+ ($87.78\%$,  0.79 F1 points above the strongest baseline), and is within $0.60$ and $0.06$ F1 points of the best method on WHU-CD and LEVIR-CD. Figure~\ref{fig:qualitative} shows the same contrast qualitatively across three change regimes on SYSU-CD: the baselines tend to either flag unchanged content or miss  genuine changes, while our predictions  track the annotated regions more closely and degrade only slightly once BSCS removes most of the FFN channels.

\subsection{Ablation Study}
\label{sec:ablation}

 We study each design choice on the category-agnostic SYSU-CD benchmark,
where adaptation matters most (Table~\ref{tab:ablation}).

  \subsubsection{Component build-up.}
  A pair-blind variant, in which each stream is adapted independently so the
  adapter never sees the other observation, improves over the frozen baseline by 0.87 F1 points. Making the adaptation pair-aware adds a further $0.97$ F1 points, more than
  the pair-blind adapter itself contributes, which indicates that the gain
  originates from pair-conditioned reshaping rather
  than from extra trainable capacity. CPGR then adds $0.51$ F1 points by reusing the
cached change responses during decoding. Additional prior ablations
in Appendix~B show that the spatial and delta responses are
individually beneficial, while their combination achieves the best
result (86.12\% F1).

  \subsubsection{What makes CGLA work.}
  We probe CGLA's two central choices at identical parameter count. Removing
  the signed temporal residual (``w/o $\Delta$ injection'') leaves only the
  gated local self-update and isolates the pairwise injection; replacing it
  with a symmetric one ($s_1{=}s_2{=}{+}1$) discards the opposite signs and
  isolates the antisymmetry. Both trail the full model, by $0.94$ and $0.64$
  F1 points, so neither the injection nor its antisymmetry is incidental.

  \subsubsection{Chunk-selection quality.}
  At the same retention budget $k{=}6$, we compare four selection rules, each
  retrained with its own retained subset so that the comparison measures the
  chunks rather than a test-time mismatch. BSCS Top-$6$ selects per batch
  from the learned scores; Static Top-$6$ keeps one globally fixed mask per
  block for all inputs; Random-$6$ keeps a fixed random subset; and Bottom-$6$
  keeps the six lowest-ranked chunks. The corresponding F1 scores are $85.29\%$, $84.58\%$,
$83.26\%$, and $82.77\%$, respectively. The $0.71$-point F1 gap between Top-$6$ and Static
  Top-$6$ isolates the value of selecting per input, and the $1.32$-point F1 gap between
  Static Top-$6$ and Random-$6$ shows that the learned ranking matters even
  when frozen; Bottom-$6$ confirms its direction.

\subsection{Accuracy--Efficiency Trade-off of BSCS}
\label{sec:efficiency}

The retention level $k$ sets the computational budget of BSCS. Table~\ref{tab:efficiency} reports the resulting trade-off on SYSU-CD, where each operating point is trained at its own budget so that the selection policy and the remaining parameters are fully adapted to it. All efficiency numbers are measured on a single NVIDIA A800 GPU with batch size 16, using 5 warm-up batches and 20 measured batches. The measured time includes the policy forward pass and the per-batch gathering of the retained submatrices. GFLOPs are estimated with the PyTorch profiler, FFN time is the total feed-forward time across all blocks amortized per pair, and speed-up is computed relative to our unpruned model.

Adaptation is inexpensive: over the frozen DINO baseline, CGLA and CPGR add 6\% GFLOPs for a 2.35-point F1 gain. BSCS more than recovers this cost: $k{=}10$ lifts throughput to 13.20 pairs/s at almost no accuracy loss, the default $k{=}6$ reaches $1.41\times$ at 85.29\% F1, and $k{=}4$ trades accuracy for $1.54\times$. Against ChangeDINO, our $k{=}6$ model is 0.59 F1 points more accurate, $1.37\times$ faster, and uses 34\% fewer GFLOPs. The FFN time locates the saving: feed-forward layers are about half of the unpruned forward pass, and at $k{=}6$ their cost falls to 17.84 ms per pair, realizing $97\%$ of the ideal  reduction. All three unpruned models share the same FFN time, confirming that CGLA and CPGR leave the feed-forward path untouched. Appendix~B further evaluates the same Top-$6$ checkpoint
with inference batch sizes from 1 to 16. The maximum F1 variation
is only $0.05$ points, indicating that the batch-shared selection
is stable across different batch sizes.

\begin{table}[!t]
  \centering
  \small
  \captionsetup{skip=2pt}
  \caption{Accuracy--efficiency trade-off on SYSU-CD. All profiled models use the same $512\times512$ backbone input
resolution and are measured under the same hardware and batch-size
setting.}
  \label{tab:efficiency}
  \setlength{\tabcolsep}{4pt}
  \renewcommand{\arraystretch}{1.05}
  \resizebox{\columnwidth}{!}{%
    \begin{tabular}{lccccc}
      \toprule
      Method & F1 & GFLOPs & FFN Time & Thr. & Speed \\
             &\%    &        & (ms/pair) & (pairs/s) & ($\times$) \\
      \midrule
      DINO baseline
        & 83.77 & 1278.66 & 45.17 & 11.38 & 1.03 \\
      Ours
        & 86.12 & 1360.83 & 45.17 & 11.00 & 1.00 \\
      ChangeDINO
        & 84.70 & 1269.10 & 45.17 & 11.32 & 1.03 \\
      \midrule
      Ours+BSCS ($k{=}10$)
        & 85.75 & 1050.15 & 29.15 & 13.20 & 1.20 \\
      Ours+BSCS ($k{=}8$)
        & 85.60 & 946.57 & 23.48 & 14.23 & 1.29 \\
      Ours+BSCS ($k{=}6$)
        & 85.29 & 842.99 & 17.84 & 15.48 & 1.41 \\
      Ours+BSCS ($k{=}4$)
        & 84.71 & 739.41 & 12.31 & 16.90 & 1.54 \\
      \bottomrule
    \end{tabular}%
  }
  \vspace{-2mm}
\end{table}
 
\subsection{Adaptation and FFN Prunability}
  \label{sec:prune-analysis}
  \begin{table}[!t]
    \centering
    \small
     \captionsetup{skip=2pt}
   \caption{FFN prunability on SYSU-CD and LEVIR-CD.
$\Delta$F1 is relative to each arm's unpruned setting.}
    \label{tab:redundancy}
    \resizebox{\columnwidth}{!}{%
    \begin{tabular}{llcccc}
      \toprule
      & & \multicolumn{2}{c}{w/o CGLA} & \multicolumn{2}{c}{w/ CGLA} \\
      \cmidrule(lr){3-4}\cmidrule(lr){5-6}
      Dataset & Retention & F1 & $\Delta$F1 & F1 & $\Delta$F1 \\
      \midrule
      \multirow{3}{*}{SYSU-CD}
        & $16/16$ (100\%) & 83.77 & --      & 85.59 & --      \\
        & $8/16$  (50\%)  & 82.88 & $-0.89$ & 85.45 & $-0.14$ \\
        & $4/16$  (25\%)  & 81.01 & $-2.76$ & 84.80 & $-0.79$ \\
      \midrule
      \multirow{3}{*}{LEVIR-CD}
        & $16/16$ (100\%) & 91.85 & --      & 92.19 & --      \\
        & $8/16$  (50\%)  & 90.98 & $-0.87$ & 91.72 & $-0.46$ \\
        & $4/16$  (25\%)  & 90.13 & $-1.72$ & 91.38 & $-0.80$ \\
      \bottomrule
    \end{tabular}}
     \vspace{-4mm}
  \end{table}

  BSCS never observes CGLA response maps; it selects FFN chunks only from pooled
  token statistics. This raises a natural question:
does adapting the backbone also make its FFN computation easier to prune? To
  test this, we disable the CPGR prior in both arms, keep the decoder fixed, and
  compare an unadapted backbone with a CGLA-adapted one under the same
  batch-shared pruning budgets.

  Table~\ref{tab:redundancy} gives a consistent answer. As retention decreases,
  the CGLA-adapted backbone loses substantially less accuracy than the
  unadapted one. On SYSU-CD, pruning to $25\%$ retention reduces the unadapted
  backbone by $2.76$ F1 points, but the adapted backbone drops by only $0.79$ F1 points. On
  LEVIR-CD, the corresponding drops are $1.72$ and $0.80$ F1 points. The advantage grows under stronger pruning, indicating that the
  task-adapted representation is more compatible with aggressive FFN chunk selection.

  The result suggests an indirect
  coupling: once CGLA reshapes the token states passed to later blocks, the
  policy operates on a feature distribution that is easier to approximate with
  fewer FFN chunks. Because the two arms are trained separately, we treat this
  as evidence of system-level compatibility rather than a strict causal
  isolation.

\section{Conclusion}
We revisited how a frozen VFM can be turned into a change detector and argued that the pair should be encoded as a change inside the backbone, not compared after it. AdaDINO realizes this with three lightweight parts. CGLA couples the two temporal streams after selected frozen blocks through a gated local update and an antisymmetric residual; BSCS keeps only a shared subset of feed-forward chunks so that inference stays dense; and CPGR reuses the change responses cached in the encoder to guide coarse-to-fine decoding. Across four benchmarks AdaDINO matches or exceeds VFM-based detectors, with the largest margin on the category-agnostic SYSU-CD. There it remains the most accurate method even after most of the feed-forward width is removed, while running faster than the baselines we profile. One limitation remains: our evidence for improved prunability rests on a matched comparison of separately trained models. Disentangling why the adapted representation is easier to prune, and extending the approach to semantic and multi-class change, are natural next steps.

\section*{Acknowledgements}
This work is supported by the National Natural Science Foundation of China (No. 62576176). The computational resources are supported by the Supercomputing Center of Nankai University (NKSC).

\bibliography{aaai2027}

@String{AAAI      = {Proceedings of the AAAI Conference on Artificial Intelligence}}

@String{CVPR      = {Proceedings of the IEEE/CVF Conference on Computer Vision and Pattern Recognition}}

@String{ICIP      = {IEEE International Conference on Image Processing}}

@String{IGARSS    = {IEEE International Geoscience and Remote Sensing Symposium}}

@String{ICLR      = {International Conference on Learning Representations}}

@String{NEURIPS   = {Advances in Neural Information Processing Systems}}

@String{TGRS      = {IEEE Transactions on Geoscience and Remote Sensing}}

@String{GRSL      = {IEEE Geoscience and Remote Sensing Letters}}

@String{JSTARS    = {IEEE Journal of Selected Topics in Applied Earth Observations and Remote Sensing}}

@String{RS        = {Remote Sensing}}

@String{ISPRSJPRS = {ISPRS Journal of Photogrammetry and Remote Sensing}}

@String{ISPRSANN  = {ISPRS Annals of the Photogrammetry, Remote Sensing and Spatial Information Sciences}}

@String{ICCV = {Proceedings of the IEEE/CVF International Conference on Computer Vision}}

@article{hussain2013change,
  author    = {Hussain, Masroor and Chen, Dongmei and Cheng, Angela and Wei, Hui and Stanley, David},
  title     = {Change Detection from Remotely Sensed Images: {F}rom Pixel-Based to Object-Based Approaches},
  journal   = ISPRSJPRS,
  volume    = {80},
  pages     = {91--106},
  year      = {2013},
  publisher = {Elsevier},
}

@article{chen2020spatial,
  author    = {Chen, Hao and Shi, Zhenwei},
  title     = {A Spatial-Temporal Attention-Based Method and a New Dataset for Remote Sensing Image Change Detection},
  journal   = RS,
  volume    = {12},
  number    = {10},
  pages     = {1662},
  year      = {2020},
  publisher = {MDPI},
}

@article{whu,
  author    = {Ji, Shunping and Wei, Shiqing and Lu, Meng},
  title     = {Fully Convolutional Networks for Multisource Building Extraction from an Open Aerial and Satellite Imagery Data Set},
  journal   = TGRS,
  volume    = {57},
  number    = {1},
  pages     = {574--586},
  year      = {2019},
  publisher = {IEEE},
}

@article{sysu,
  author    = {Shi, Qian and Liu, Mengxi and Li, Shengchen and Liu, Xiaoping and Wang, Fei and Zhang, Liangpei},
  title     = {A Deeply Supervised Attention Metric-Based Network and an Open Aerial Image Dataset for Remote Sensing Change Detection},
  journal   = TGRS,
  volume    = {60},
  pages     = {1--16},
  year      = {2022},
  publisher = {IEEE},
}

@article{s2looking,
  author    = {Shen, Li and Lu, Yao and Chen, Hao and Wei, Hao and Xie, Donghai and Yue, Jiabao and Chen, Rui and Lv, Shouye and Jiang, Bitao},
  title     = {{S2Looking}: {A} Satellite Side-Looking Dataset for Building Change Detection},
  journal   = RS,
  volume    = {13},
  number    = {24},
  pages     = {5094},
  year      = {2021},
  publisher = {MDPI},
}

@inproceedings{fcsiam,
  author    = {Daudt, Rodrigo Caye and Le Saux, Bertrand and Boulch, Alexandre},
  title     = {Fully Convolutional Siamese Networks for Change Detection},
  booktitle = ICIP,
  pages     = {4063--4067},
  year      = {2018},
}

@article{bit,
  author    = {Chen, Hao and Qi, Zipeng and Shi, Zhenwei},
  title     = {Remote Sensing Image Change Detection with Transformers},
  journal   = TGRS,
  volume    = {60},
  pages     = {1--14},
  year      = {2022},
  publisher = {IEEE},
}

@inproceedings{changeformer,
  author    = {Bandara, Wele Gedara Chaminda and Patel, Vishal M.},
  title     = {A Transformer-Based Siamese Network for Change Detection},
  booktitle = IGARSS,
  pages     = {207--210},
  year      = {2022},
}

@article{changer,
  author    = {Fang, Sheng and Li, Kaiyu and Li, Zhe},
  title     = {{Changer}: {F}eature Interaction Is What You Need for Change Detection},
  journal   = TGRS,
  volume    = {61},
  pages     = {1--11},
  year      = {2023},
  publisher = {IEEE}
}

@inproceedings{sarasnet,
  author    = {Chen, Chao-Peng and Hsieh, Jun-Wei and Chen, Ping-Yang and Hsieh, Yi-Kuan and Wang, Bor-Shiun},
  title     = {{SARAS-Net}: {S}cale and Relation Aware Siamese Network for Change Detection},
  booktitle = AAAI,
  pages     = {14187--14195},
  year      = {2023}
}

@inproceedings{ttp,
  author    = {Chen, Keyan and Liu, Chengyang and Li, Wenyuan and Liu, Zili and Chen, Hao and Zhang, Haotian and Zou, Zhengxia and Shi, Zhenwei},
  title     = {{Time Travelling Pixels}: {B}itemporal Features Integration with Foundation Model for Remote Sensing Image Change Detection},
  booktitle = IGARSS,
  pages     = {8581--8584},
  year      = {2024},
}

@article{cdmamba,
  author    = {Zhang, Haotian and Chen, Keyan and Liu, Chenyang and Chen, Hao and Zou, Zhengxia and Shi, Zhenwei},
  title     = {{CDMamba}: {I}ncorporating Local Clues into Mamba for Remote Sensing Image Binary Change Detection},
  journal   = TGRS,
  volume    = {63},
  pages     = {1--16},
  year      = {2025},
  publisher = {IEEE},
}

@article{stmamba,
  author    = {Zhao, Jiaqi and Xie, Jianpeng and Zhou, Yong and Du, Wen-Liang and Yao, Rui and El Saddik, Abdulmotaleb},
  title     = {{ST-Mamba}: {S}patio-Temporal Synergistic Model for Remote Sensing Change Detection},
  journal   = TGRS,
  volume    = {63},
  pages     = {1--13},
  year      = {2025},
  publisher = {IEEE}
}

@article{snunet,
  author    = {Fang, Sheng and Li, Kaiyu and Shao, Jinyuan and Li, Zhe},
  title     = {{SNUNet-CD}: {A} Densely Connected Siamese Network for Change Detection of {VHR} Images},
  journal   = GRSL,
  volume    = {19},
  pages     = {1--5},
  year      = {2022},
  publisher = {IEEE}
}

@article{mambacd,
  author    = {Chen, Hongruixuan and Song, Jian and Han, Chengxi and Xia, Junshi and Yokoya, Naoto},
  title     = {{ChangeMamba}: {R}emote Sensing Change Detection with Spatiotemporal State Space Model},
  journal   = TGRS,
  volume    = {62},
  pages     = {1--20},
  year      = {2024},
  publisher = {IEEE}
}

@article{rsmamba,
  author    = {Zhao, Sijie and Chen, Hao and Zhang, Xueliang and Xiao, Pengfeng and Bai, Lei and Ouyang, Wanli},
  title     = {{RS-Mamba} for Large Remote Sensing Image Dense Prediction},
  journal   = TGRS,
  volume    = {62},
  pages     = {1--14},
  year      = {2024},
  publisher = {IEEE}
}

@article{changetitans,
  author    = {Yang, Zhenyu and Pei, Gensheng and Yao, Yazhou and Zhou, Tianfei and Ding, Lizhong and Shen, Fumin},
  title     = {{ChangeTitans}: {T}oward Remote Sensing Change Detection with Neural Memory},
  journal   = TGRS,
  volume    = {63},
  pages     = {1--14},
  year      = {2025},
  publisher = {IEEE}
}

@article{ban,
  author    = {Li, Kaiyu and Cao, Xiangyong and Meng, Deyu},
  title     = {A New Learning Paradigm for Foundation Model-Based Remote-Sensing Change Detection},
  journal   = TGRS,
  volume    = {62},
  pages     = {1--12},
  year      = {2024},
  publisher = {IEEE}
}

@article{changedino,
  author    = {Cheng, Ching-Heng and Hsu, Chih Chung},
  title     = {{ChangeDINO}: {DINOv3}-Driven Building Change Detection in Optical Remote Sensing Imagery},
  journal   = ISPRSANN,
  volume    = {XI-3-2026},
  pages     = {523--530},
  year      = {2026},
  publisher = {Copernicus Publications}
}

@article{peftcd,
  author    = {Dong, Sijun and Hu, Yuxuan and Wang, Libo and Chen, Geng and Meng, Xiaoliang},
  title     = {{PeftCD}: {L}everaging Vision Foundation Models with Parameter-Efficient Fine-Tuning for Remote Sensing Change Detection},
  journal   = JSTARS,
  pages     = {1--16},
  year      = {2026},
  publisher = {IEEE}
}

@article{faewnet,
  author    = {Li, Yun-Cheng and Lei, Sen and Zhao, Yi-Tao and Li, Heng-Chao and Li, Jun and Plaza, Antonio},
  title     = {{SAM}-Based Building Change Detection with Distribution-Aware Fourier Adaptation and Edge-Constrained Warping},
  journal   = TGRS,
  volume    = {63},
  pages     = {1--14},
  year      = {2025},
  publisher = {IEEE}
}

@inproceedings{adavit,
  author    = {Meng, Lingchen and Li, Hengduo and Chen, Bor-Chun and Lan, Shiyi and Wu, Zuxuan and Jiang, Yu-Gang and Lim, Ser-Nam},
  title     = {{AdaViT}: {A}daptive Vision Transformers for Efficient Image Recognition},
  booktitle = CVPR,
  pages     = {12309--12318},
  year      = {2022}
}

@inproceedings{dynamicvit,
  author    = {Rao, Yongming and Zhao, Wenliang and Liu, Benlin and Lu, Jiwen and Zhou, Jie and Hsieh, Cho-Jui},
  title     = {{DynamicViT}: {E}fficient Vision Transformers with Dynamic Token Sparsification},
  booktitle = NEURIPS,
  pages     = {13937--13949},
  year      = {2021}
}

@inproceedings{avit,
  author    = {Yin, Hongxu and Vahdat, Arash and Alvarez, Jose M. and Mallya, Arun and Kautz, Jan and Molchanov, Pavlo},
  title     = {{A-ViT}: {A}daptive Tokens for Efficient Vision Transformer},
  booktitle = CVPR,
  pages     = {10809--10818},
  year      = {2022}
}

@inproceedings{lora,
  author    = {Hu, Edward J. and Shen, Yelong and Wallis, Phillip and Allen-Zhu, Zeyuan and Li, Yuanzhi and Wang, Shean and Wang, Lu and Chen, Weizhu},
  title     = {{LoRA}: {L}ow-Rank Adaptation of Large Language Models},
  booktitle = ICLR,
  pages     = {1--16},
  year      = {2022}
}

@inproceedings{vitadapter,
  author    = {Chen, Zhe and Duan, Yuchen and Wang, Wenhai and He, Junjun and Lu, Tong and Dai, Jifeng and Qiao, Yu},
  title     = {Vision Transformer Adapter for Dense Predictions},
  booktitle = ICLR,
  pages     = {1--14},
  year      = {2023}
}

@inproceedings{tome,
  author    = {Bolya, Daniel and Fu, Cheng-Yang and Dai, Xiaoliang and Zhang, Peizhao and Feichtenhofer, Christoph and Hoffman, Judy},
  title     = {Token Merging: {Y}our {ViT} But Faster},
  booktitle = ICLR,
  pages     = {1--12},
  year      = {2023}
}

@misc{amp,
  author        = {Shen, Chengchao},
  title         = {Adaptive {MLP} Pruning for Large Vision Transformers},
  year          = {2026},
  eprint        = {2603.08100},
  archivePrefix = {arXiv},
  primaryClass  = {cs.CV},
  url           = {https://arxiv.org/abs/2603.08100},
}

@article{oaip,
  author    = {Zhang, Chengyang and Wang, Xueqian and Li, Weiming and Li, Gang and Song, Huina and Song, Zhaohui and Zhao, Jie and Plaza, Antonio},
  title     = {Lightweight Change Detection in Heterogeneous Remote Sensing Images with Online All-Integer Pruning Training},
  journal   = JSTARS,
  volume    = {19},
  pages     = {8719--8739},
  year      = {2026},
  publisher = {IEEE}
}

@article{lcdnet,
  author    = {Liu, Wenyu and Li, Jindong and Wang, Haoji and Tan, Run and Fu, Yali and Tian, Qichuan},
  title     = {{LCD-Net}: {A} Lightweight Remote Sensing Change Detection Network Combining Feature Fusion and Gating Mechanism},
  journal   = JSTARS,
  volume    = {18},
  pages     = {7769--7780},
  year      = {2025},
  publisher = {IEEE}
}

@misc{dynamicvis,
  author        = {Chen, Keyan and Liu, Chenyang and Chen, Bowen and Li, Wenyuan and Zou, Zhengxia and Shi, Zhenwei},
  title         = {{DynamicVis}: {A}n Efficient and General Visual Foundation Model for Remote Sensing Image Understanding},
  year          = {2025},
  eprint        = {2503.16426},
  archivePrefix = {arXiv},
  primaryClass  = {cs.CV},
  url           = {https://arxiv.org/abs/2503.16426}
}

@inproceedings{feaspect,
  author    = {Zang, Qi and Zhao, Dong and Wang, Shuang and Quan, Dou and Zhong, Zhun},
  title     = {Feature Spectrum Learning for Remote Sensing Change Detection},
  booktitle = CVPR,
  pages     = {12647--12657},
  year      = {2025}
}

@inproceedings{gencd,
  author    = {Zhang, Mingwei and Hu, Jingtao and Li, Qiang and Wang, Qi},
  title     = {Exploring Generalizable Remote Sensing Change Detection via
               Low-Rank Exchange Adaptation of Vision Foundation Model},
  booktitle = AAAI,
  pages     = {12663--12671},
  year      = {2026}
}

@inproceedings{peftvit,
  author    = {He, Xuehai and Li, Chunyuan and Zhang, Pengchuan and
               Yang, Jianwei and Wang, Xin Eric},
  title     = {Parameter-Efficient Model Adaptation for Vision Transformers},
  booktitle = AAAI,
  pages     = {817--825},
  year      = {2023}
}

@inproceedings{lorand,
  author    = {Yin, Dongshuo and Yang, Yiran and Wang, Zhechao and
               Yu, Hongfeng and Wei, Kaiwen and Sun, Xian},
  title     = {{1\% VS 100\%}: {P}arameter-Efficient Low Rank Adapter
               for Dense Predictions},
  booktitle = CVPR,
  pages     = {20116--20126},
  year      = {2023}
}

@inproceedings{sparsevit,
  author    = {Chen, Xuanyao and Liu, Zhijian and Tang, Haotian and
               Yi, Li and Zhao, Hang and Han, Song},
  title     = {{SparseViT}: {R}evisiting Activation Sparsity for Efficient
               High-Resolution Vision Transformer},
  booktitle = CVPR,
  pages     = {2061--2070},
  year      = {2023}
}

@inproceedings{sam,
  author    = {Kirillov, Alexander and Mintun, Eric and Ravi, Nikhila and
               Mao, Hanzi and Rolland, Chloe and Gustafson, Laura and
               Xiao, Tete and Whitehead, Spencer and Berg, Alexander C. and
               Lo, Wan-Yen and Dollar, Piotr and Girshick, Ross},
  title     = {Segment Anything},
  booktitle = ICCV,
  pages     = {4015--4026},
  year      = {2023}
}

@misc{dinov3,
  author        = {Sim{\'e}oni, Oriane and Vo, Huy V. and
                   Seitzer, Maximilian and Baldassarre, Federico and
                   Oquab, Maxime and Jose, Cijo and Khalidov, Vasil and
                   Szafraniec, Marc and Yi, Seungeun and
                   Ramamonjisoa, Micha{\"e}l and Massa, Francisco and
                   Haziza, Daniel and Wehrstedt, Luca and Wang, Jianyuan and
                   Darcet, Timoth{\'e}e and Moutakanni, Th{\'e}o and
                   Sentana, Leonel and Roberts, Claire and Vedaldi, Andrea and
                   Tolan, Jamie and Brandt, John and Couprie, Camille and
                   Mairal, Julien and J{\'e}gou, Herv{\'e} and
                   Labatut, Patrick and Bojanowski, Piotr},
  title         = {{DINOv3}},
  year          = {2025},
  eprint        = {2508.10104},
  archivePrefix = {arXiv},
  primaryClass  = {cs.CV},
  url           = {https://arxiv.org/abs/2508.10104},
}
\clearpage
\appendix
\setcounter{secnumdepth}{2}

% Appendix numbering.
\setcounter{table}{0}
\setcounter{figure}{0}
\setcounter{equation}{0}
\renewcommand{\thetable}{S\arabic{table}}
\renewcommand{\thefigure}{S\arabic{figure}}
\renewcommand{\theequation}{S\arabic{equation}}

\section{Additional Implementation Details}
\label{app:implementation}

\subsection{Training Objective}
\label{app:training-objective}
Let $\widehat{\mathbf P}$ denote the final refined prediction and
$\{\mathbf P_r\}_{r=2}^{5}$ the four pyramid predictions, all upsampled to
the ground-truth resolution before supervision. The complete training
objective is
\begin{equation}
\begin{aligned}
\mathcal L_{\mathrm{total}} ={}&
\tfrac{1}{2}\mathcal L_{\mathrm{foc}}
  (\widehat{\mathbf P},\mathbf Y)
+\mathcal L_{\mathrm{dice}}
  (\widehat{\mathbf P},\mathbf Y) \\
&+\tfrac{1}{2}\sum_{r=2}^{5}
  \mathcal L_{\mathrm{foc}}(\mathbf P_r,\mathbf Y) \\
&+\tfrac{1}{2}\sum_{r=2}^{5}
  \mathcal L_{\mathrm{dice}}(\mathbf P_r,\mathbf Y) \\
&+\lambda_{\mathrm{cgla}}(e)\mathcal L_{\mathrm{cgla}}.
\end{aligned}
\label{eq:total-loss}
\end{equation}
The focal loss operates on two-class logits with $\alpha=0.25$ and
$\gamma=4$. The Dice term is the multi-class micro Dice loss. The CGLA
auxiliary term supervises the channel-averaged absolute delta response
$q_{\Delta}^{l}$ at each adapted depth. With the resized change mask
$M^{l}$ and margin $m=0.1$, it is
\begin{align}
\mathcal L_{\mathrm{cgla}}
&=\frac{1}{|\mathcal L_c|}
  \sum_{l\in\mathcal L_c}
  \left(\mathcal L_{\mathrm{chg}}^{l}
       +\mathcal L_{\mathrm{unchg}}^{l}\right),
\label{eq:cgla-loss}\\
\mathcal L_{\mathrm{chg}}^{l}
&=\frac{\sum M^{l}
  [\max(0,m-q_{\Delta}^{l})]^2}
  {\max(\sum M^{l},1)},\\
\mathcal L_{\mathrm{unchg}}^{l}
&=\frac{\sum (1-M^{l})(q_{\Delta}^{l})^2}
  {\max(\sum (1-M^{l}),1)}.
\end{align}
The coefficient $\lambda_{\mathrm{cgla}}(e)$ is zero during the first five
epochs and then increases linearly over ten epochs to $0.01$.

\subsection{Architecture Specifications}
\label{app:architecture}
Table~\ref{tab:architecture} lists implementation-specific settings omitted
from the main paper. The frozen backbone is DINOv3 ViT-L/16 with embedding
width $D=1024$.

\begin{table*}[!t]
  \centering
  \footnotesize
  \captionsetup{skip=2pt}
  \caption{Implementation-specific architecture settings of AdaDINO.}
  \label{tab:architecture}
  \setlength{\tabcolsep}{5pt}
  \renewcommand{\arraystretch}{1.03}
  \begin{tabular}{@{}p{0.12\textwidth}p{0.19\textwidth}p{0.62\textwidth}@{}}
    \toprule
    Component & Item & Configuration \\
    \midrule
    \multirow{8}{*}{CGLA}
      & Insertion depths & $\{2,5,8,11,14,17,20,23\}$ \\
      & Hidden width & $H=0.5D=512$ \\
      & $f_{\mathrm{rel}}$ & $1\!\times\!1$ Conv $(4D\!\rightarrow\!H)$, GN(8), SiLU \\
      & $f_{\mathrm{ctx}}$ & depthwise $3\!\times\!3$ Conv, GN(8), SiLU \\
      & $f_{\Delta}$ & $1\!\times\!1$ Conv, GN, SiLU, DWConv $3\!\times\!3$, GN, SiLU, $1\!\times\!1$ Conv to $D$ \\
      & $f_{\mathrm{loc}}$ & shared DWConv $3\!\times\!3$, GELU, $1\!\times\!1$ Conv \\
      & Scale initialization & $\eta=0.05$, $\bm\gamma_{\Delta}=0.05$, $\bm\gamma_{\mathrm{self}}=0$ \\
      & Trainable parameters & 4.48M per module; 35.88M over eight modules \\
    \midrule
    \multirow{6}{*}{BSCS}
      & Applied blocks & all 24 DINO blocks \\
      & FFN partition & width 4096, $C=16$, 256 channels per chunk \\
      & Policy input & pooled $[\mathbf x_1,\mathbf x_2,|\mathbf x_1-\mathbf x_2|,\mathbf x_1\odot\mathbf x_2]$ \\
      & Policy MLP & Linear$(4096,256)$--GELU--Linear$(256,16)$ \\
      & Selection settings & $\tau=1$, three-epoch warm-up, default $k=6$ \\
      & Trainable parameters & 1.05M per block; 25.27M over 24 blocks \\
    \midrule
    \multirow{6}{*}{CPGR}
      & Input features & four bi-temporal scales, 256 channels per stream and scale \\
      & Prior mode & spatial and delta responses; local response unused \\
      & Depth grouping & $\{2,5\}\!\rightarrow\!P_2$, $\{8,11\}\!\rightarrow\!P_3$, $\{14,17\}\!\rightarrow\!P_4$, $\{20,23\}\!\rightarrow\!P_5$; $P_5$ prior is not injected \\
      & Decoder width & 256 channels at all stages \\
      & Prior/refiner settings & prior scale initialized to 0.05; soft opening $3\!\times\!3$ and closing $5\!\times\!5$ \\
      & Trainable parameters & 9.41M including the final morphological refiner \\
    \bottomrule
  \end{tabular}
\end{table*}

\subsection{Training Configuration}
We use AdamW with betas $(0.9,0.999)$, without gradient clipping or mixed
precision. The physical batch size is eight image pairs and gradients are
accumulated over two steps, yielding an effective batch size of 16. Input
patches are prepared at $256\times256$ and resized inside the encoder to
$512\times512$ using bilinear interpolation with
\texttt{align\_corners=True} and antialiasing. Images are normalized with
mean $(0.430,0.411,0.296)$ and standard deviation
$(0.213,0.156,0.143)$.

Training augmentation includes temporal swapping, horizontal flipping,
vertical flipping, right-angle rotation, color jitter, and random resized
cropping, each with probability $0.5$. Rotation angles are sampled from
$\{90^{\circ},180^{\circ},270^{\circ}\}$. Color-jitter brightness,
contrast, and saturation factors are independently sampled from
$[0.75,1.25]$ and the same jitter is applied to both temporal images. Random
resized cropping outputs $256\times256$ patches with scale
$[0.333,1.0]$ and aspect ratio $[0.75,1.333]$; bilinear interpolation is
used for images and nearest-neighbor interpolation for labels.

\section{Additional Ablation Studies}
\label{app:ablations}

\subsection{CPGR Prior Ablation}
\label{app:cpgr-prior}
We evaluate the spatial and delta CGLA responses used to guide
decoder refinement. As shown in Table~\ref{tab:cpgr-prior}, each
response improves over the CGLA-only model without prior-guided
refinement, and their combination performs best.

\begin{table}[H]
  \centering
  \small
  \captionsetup{skip=2pt}
  \caption{Ablation of CGLA priors used by CPGR on SYSU-CD (F1/IoU, \%).}
  \label{tab:cpgr-prior}
  \setlength{\tabcolsep}{6pt}
  \renewcommand{\arraystretch}{1.08}
  \begin{tabular}{lcc}
    \toprule
    Prior configuration & F1 & IoU \\
    \midrule
     CGLA only  & 85.61 & 74.84 \\
    Spatial only & 85.94 & 75.35 \\
    Delta only & 86.01 & 75.45 \\
    Spatial + delta & 86.12 & 75.62 \\
    \bottomrule
  \end{tabular}
\end{table}

\subsection{BSCS Stability across Inference Batch Sizes}
\label{app:batch-size}
Table~\ref{tab:batch-size} evaluates the same Top-$6$ checkpoint under four
inference batch sizes on the SYSU-CD test set. Accuracy is nearly unchanged:
the maximum F1 spread is only 0.05 points. This indicates that the
batch-shared selection is not strongly dependent on a particular inference
batch size. Throughput increases with batch size and begins to saturate near
batch size 16. 

\begin{table}[H]
  \centering
  \small
  \captionsetup{skip=2pt}
  \caption{BSCS batch-size stability on SYSU-CD (F1/IoU, \%), measured on
  one NVIDIA A800 GPU. Throughput is measured in bi-temporal pairs per
  second.}
  \label{tab:batch-size}
  \setlength{\tabcolsep}{6pt}
  \renewcommand{\arraystretch}{1.08}
  \begin{tabular}{rccc}
    \toprule
    Batch size & F1 & IoU & pairs/s \\
    \midrule
     1 & 85.24 & 74.28 & 7.33 \\
     4 & 85.27 & 74.32 & 12.30 \\
     8 & 85.25 & 74.29 & 13.91 \\
    16 & 85.29 & 74.35 & 15.48 \\
    \bottomrule
  \end{tabular}
\end{table}

\subsection{BSCS Stability across Batch Compositions}
\label{app:batch-composition}
Because BSCS uses one chunk mask shared across a batch, the mask
applied to a given pair can vary with the composition of that batch.
We therefore re-evaluate the same Top-$6$ checkpoint under five random
orderings of the SYSU-CD test set at a fixed inference batch size of 16. As reported in Table~\ref{tab:robustness}, F1 stays within a range of
$0.04$ points across the five orderings, and even the least favorable
ordering differs from the default evaluation by less than $0.1$ points.
Together with Table~\ref{tab:batch-size}, this indicates that the
batch-shared selection is insensitive both to the inference batch size and to
how the test set is partitioned into batches.
\begin{table}[H]
  \centering
  \small
  \captionsetup{skip=2pt}
  \caption{Robustness of the Top-$6$ checkpoint to test-set ordering and to
  temporal order on SYSU-CD (F1/IoU, \%), evaluated at inference batch
  size 16.}
  \label{tab:robustness}
  \setlength{\tabcolsep}{6pt}
  \renewcommand{\arraystretch}{1.08}
  \begin{tabular}{lccc}
    \toprule
    Test setting & F1 & IoU & $\Delta$F1 \\
    \midrule
    Default ordering & 85.29 & 74.35 & 0 \\
    Shuffled ordering, mean of five & 85.23 & 74.26 & $-0.06$ \\
    Shuffled ordering, worst of five & 85.20 & 74.22 & $-0.09$ \\
    Temporal swap & 85.15 & 74.14 & $-0.14$ \\
    \bottomrule
  \end{tabular}
\end{table}
\subsection{Sensitivity to Temporal Order}
\label{app:temporal-order}
Because the pairwise relation descriptor uses an ordered concatenation
of the two appearance features, the resulting model is not strictly
invariant to temporal order by construction. To quantify the residual
asymmetry, we evaluate the same Top-$6$ checkpoint after exchanging
$I_1$ and $I_2$ at test time, while keeping all other settings fixed.
F1 decreases by 0.14 points (Table~\ref{tab:robustness}). The effect is small in absolute terms, although larger than the
variation across batch compositions reported above. This is consistent
with temporal swapping being encouraged through data augmentation
rather than enforced by the architecture.

\end{document}